\documentclass[letterpaper, 10 pt, conference]{ieeeconf}  

\IEEEoverridecommandlockouts                              

\usepackage{times}
\usepackage[pdftex]{graphicx}
\usepackage{graphics}
\usepackage[cmex10]{amsmath}
\usepackage{url}
\usepackage{array}
\usepackage{tabularx}
\usepackage{multirow}
\usepackage{multicol}
\usepackage{booktabs}
\usepackage{epstopdf}
\usepackage{rotating}
\usepackage{csquotes}

\usepackage{amssymb}
\usepackage{amsmath}
\usepackage[all]{xy}    
\usepackage{ifthen}
\usepackage[usenames,dvipsnames]{color}
\usepackage{enumerate}
\usepackage{bm}
\usepackage{hyperref}
\usepackage{siunitx}
\usepackage{xargs} 
\usepackage[pdftex,dvipsnames]{xcolor}  
\usepackage[colorinlistoftodos,prependcaption,textsize=tiny]{todonotes}
\newcommandx{\unsure}[2][1=]{\todo[linecolor=red,backgroundcolor=red!25,bordercolor=red,#1]{#2}}
\newcommandx{\change}[2][1=]{\todo[linecolor=blue,backgroundcolor=white!25,bordercolor=blue,fancyline,#1]{#2}}
\newcommandx{\info}[2][1=]{\todo[linecolor=OliveGreen,backgroundcolor=OliveGreen!25,bordercolor=OliveGreen,#1]{#2}}
\newcommandx{\improvement}[2][1=]{\todo[linecolor=Plum,backgroundcolor=Plum!25,bordercolor=Plum,#1]{#2}}
\newcommandx{\thiswillnotshow}[2][1=]{\todo[disable,#1]{#2}}

\makeatletter
\let\NAT@parse\undefined
\makeatother
\usepackage[numbers]{natbib}

\newcommand{\secref}[1]{Section~\ref{#1}}

\newcommand{\figref}[1]{Figure~\ref{#1}}

\newcommand*{\Cdot}{\raisebox{-0.25ex}{\scalebox{1.75}{$\cdot$}}}
\newcommand{\myparagraph}[1]{\vspace{0.1in}\noindent\textbf{#1}}

\newboolean{draft-mode}
\setboolean{draft-mode}{false}
\newcommand{\sidenote}[1]{\ifthenelse{\boolean{draft-mode}}{\marginpar{\tiny\raggedright\textsf{\hspace{0pt}#1}}}{}}
\DeclareRobustCommand{\pynote}[1]{\ifthenelse{\boolean{draft-mode}}{\textcolor{green}{\textbf{PY: #1}}}{}}
\DeclareRobustCommand{\arnote}[1]{\ifthenelse{\boolean{draft-mode}}{\textcolor{blue}{\textbf{AR: #1}}}{}}
\DeclareRobustCommand{\nfnote}[1]{\ifthenelse{\boolean{draft-mode}}{\textcolor{red}{\textbf{NF: #1}}}{}}
\DeclareRobustCommand{\mbnote}[1]{\ifthenelse{\boolean{draft-mode}}{\textcolor{cyan}{\textbf{MB: #1}}}{}}

\title{\LARGE \bf
Multi-view Self-supervised Deep Learning for 6D Pose Estimation \\in the Amazon Picking Challenge}

\author{Andy Zeng$^{1}$  Kuan-Ting Yu$^{2}$  Shuran Song$^{1}$  Daniel Suo$^{1}$  Ed Walker Jr.$^{3}$ Alberto Rodriguez$^{2}$  Jianxiong Xiao$^{4}$
\vspace{0.1cm} \\ 
$^{1}$Princeton University\quad\quad
$^{2}$Massachusetts Institute of Technology\quad\quad
$^{3}$Google\quad\quad
$^{4}$AutoX
\thanks{The authors would like to thank the MIT-Princeton APC team members for their contribution to this project, and ABB Inc. for hardware and technical support. This project is also supported by the Google Faculty Award and Intel Gift Fund to Jianxiong Xiao. Andy Zeng and Daniel Suo are supported by the Gordon Y.S. Wu fellowship. Shuran Song is supported by the Facebook fellowship. Kuan-Ting Yu is supported by award [NSF-IIS-1427050] through the National Robotics Initiative. Alberto Rodriguez is supported by the Walter Henry Gale (1929) Career Development Professorship.}
}

\begin{document}

\maketitle
\thispagestyle{empty}
\pagestyle{empty}

\begin{abstract}

Robot warehouse automation has attracted significant interest in recent years, perhaps most visibly in the Amazon Picking Challenge (APC) \cite{apcwebsite}. 
A fully autonomous warehouse pick-and-place system requires robust vision that reliably recognizes and locates objects amid cluttered environments, self-occlusions, sensor noise, and a large variety of objects.
In this paper we present an approach that leverages multi-view RGB-D data and self-supervised, data-driven learning to overcome those difficulties.
The approach was part of the MIT-Princeton Team system that took 3rd- and 4th- place in the stowing and picking tasks, respectively at APC 2016.

In the proposed approach, we segment and label multiple views of a scene with a fully convolutional neural network, and then fit pre-scanned 3D object models to the resulting segmentation to get the 6D object pose.
Training a deep neural network for segmentation typically requires a large amount of training data.
We propose a self-supervised method to generate a large labeled dataset without tedious manual segmentation.
%
We demonstrate that our system can reliably estimate the 6D pose of objects under a variety of scenarios. All code, data, and benchmarks are available at \href{http://apc.cs.princeton.edu/}{http://apc.cs.princeton.edu/}

\end{abstract}


\section{Introduction}
The last two decades have seen a rapid increase in warehouse automation technologies, satisfying the growing demand of e-commerce and providing faster, cheaper delivery. Some tasks, especially those involving physical interaction, are still hard to automate. Amazon, in collaboration with the academic community, has led a recent effort to define two such tasks: 
1) \emph{picking} an instance of a given a product ID out of a populated shelf and place it into a tote; 
and 2) \emph{stowing} a tote full of products into a populated shelf.

In this paper we describe the vision system of the MIT-Princeton Team, that took 3rd place in the stowing task and 4th in the picking task at the 2016 Amazon Picking Challenge (APC), and provide experiments to validate our design decisions. 
Our vision algorithm estimates the 6D poses of objects robustly under challenging scenarios:

\begin{itemize}
\item[$\Cdot$] \textbf{Cluttered environments}: shelves and totes may have multiple objects and could be arranged as to deceive vision algorithms (e.g., objects on top of one another).

\item[$\Cdot$] \textbf{Self-occlusion}: due to limited camera positions, the system only sees a partial view of an object.

\item[$\Cdot$] \textbf{Missing data}: commercial depth sensors are unreliable at capturing reflective, transparent, or meshed surfaces, all common in product packaging.

\item[$\Cdot$] \textbf{Small or deformable objects}: small objects provide fewer data points, while deformable objects are difficult to align to prior models.

\item[$\Cdot$] \textbf{Speed}: the total time dedicated to capturing and processing visual information is under 20 seconds.
\end{itemize}

\begin{figure}
\centering
  \includegraphics[width=8.5cm]{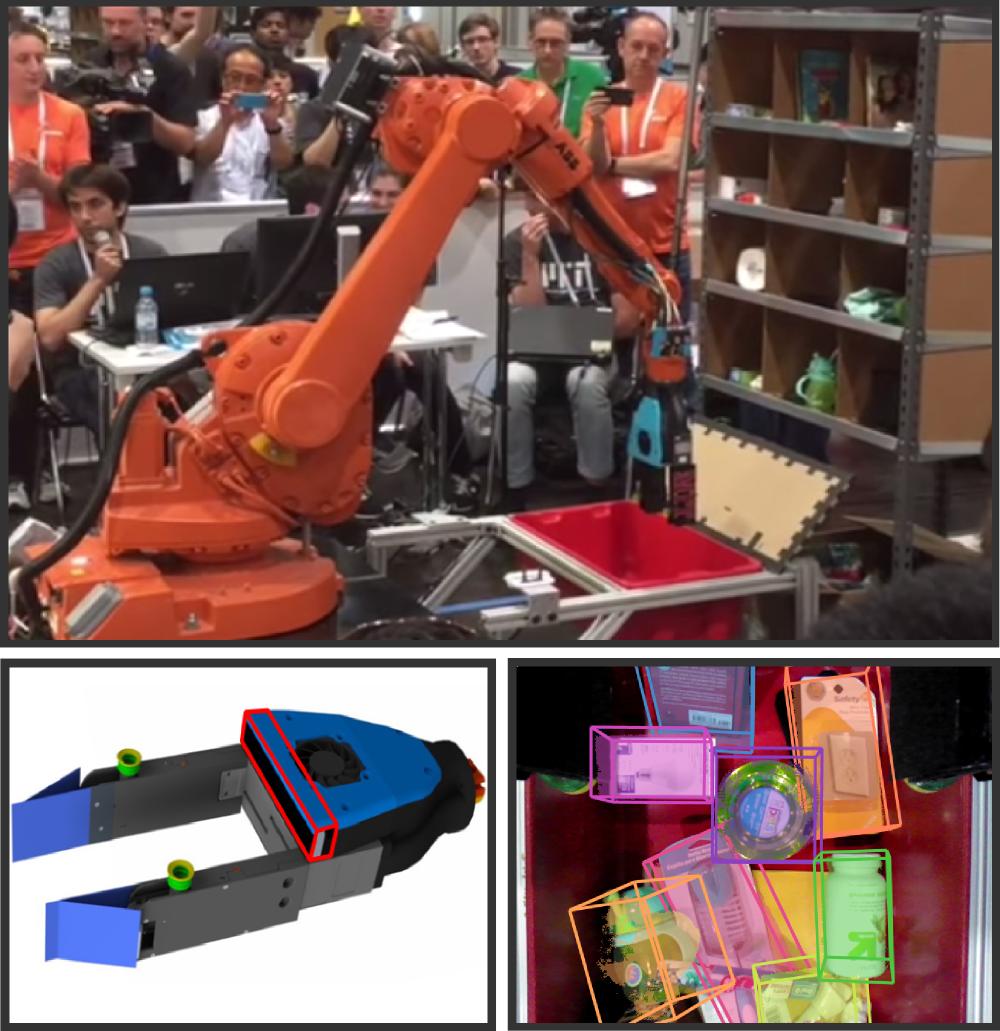}
  \caption{Top: The MIT-Princeton robotic picking system. Bottom-left: The gripper mounted with an Intel RealSense camera (outlined in red). Bottom-right: Predicted 6D object poses from our vision system during the stow-task finals of the APC 2016. Each prediction is highlighted with a colored 3D bounding box.}
  \label{fig:teaser}
\end{figure}

Our approach makes careful use of known constraints in the task---the list of possible objects and the expected background.
The algorithm first segments the object from a scene by feeding multiple-view images to a deep neural network and then fits a 3D model to a segmented point cloud to recover the object's 6D pose.
The 
deep neural network provides speed, and in combination with a multiple-view approach boosts performance in challenging scenarios.

\begin{figure*}[t]
\vspace{2mm}
\centering
  \includegraphics[width=\textwidth]{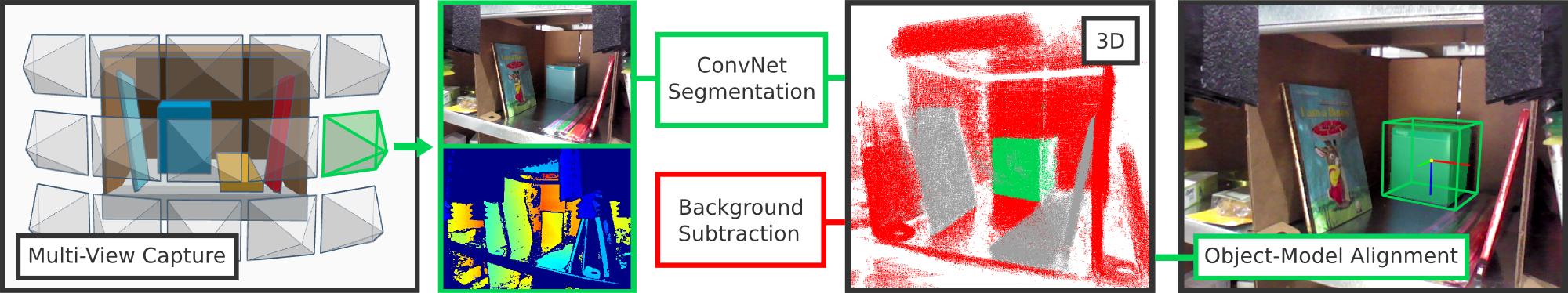}
  \caption{Overview of the vision algorithm. The robot captures color and depth images from 15 to 18 viewpoints of the scene. Each color image is fed into a fully convolutional network \cite{long2015fully} for 2D object segmentation. The result is integrated in 3D. The point cloud will then go through background removal and then aligned with a pre-scanned 3D model to obtain its 6D pose.}
  \label{fig:overview}
\end{figure*}

Training a deep neural network for segmentation requires a large amount of labeled training data. We have developed a self-supervised training procedure that automatically generated 130,000 images with pixel-wise category labels of the 39 objects in the APC. For evaluation, we constructed a testing dataset of over 7,000 manually-labeled images. 

In summary, this paper contributes with:
\begin{itemize}
\item[$\Cdot$] A robust multi-view vision system to estimate the 6D pose of objects;
\item[$\Cdot$] A self-supervised method that trains deep networks by automatically labeling training data; 
\item[$\Cdot$] A benchmark dataset for estimating object poses. \end{itemize}
All code, data, and benchmarks are publicly available~ \cite{projectwebsite}.

\section{Related Work}
\label{sec:RelatedWork}
Vision algorithms for robotic manipulation typically output 2D bounding boxes, pixel-level segmentation \cite{jonschkowski2016probabilistic, rbolessons}, or 6D poses \cite{zhang2016dorapicker, yuapc2015} of the objects.
The choice depends primarily on manipulation needs. For example, a suction based picker might have sufficient information with a 2D bounding box or with a pixel-level segmentation of the object, while a grasper might require its 6D pose. 

\myparagraph{Object segmentation.}
While the 2015 APC winning team used a histogram backprojection method~\cite{Ming-Yu} with manually defined features~\cite{rbolessons, jonschkowski2016probabilistic}, recent work in computer vision has shown that deep learning considerably improves object segmentation~\cite{long2015fully}.
In this work, we extend the state-of-the-art deep learning architecture used for image segmentation to incorporate depth and multi-view information. 

\myparagraph{Pose estimation.}
There are two primary approaches for estimating the 6D pose of an object. The first aligns 3D CAD models to 3D point clouds
with algorithms such as iterative closest point~\cite{besl1992method}. 
The second uses more elaborated local descriptors such as SIFT keypoints~\cite{SIFT} for color data or 3DMatch~\cite{3dmatch} for 3D data.
The former approach is mainly used with depth-only sensors, in scenarios where lighting changes significantly, or on textureless objects. Highly textured and rigid objects, on the other hand, benefit from local descriptors. 
Existing frameworks such as LINEMOD~\cite{LINEMOD} or MOPED~\cite{collet2011moped} work well under certain assumptions such as objects sitting on a table top with good illumination, but underperform when confronted with the limited visibility, shadows, and clutter imposed by the APC scenario~\cite{Correll2016}.

%





\myparagraph{Benchmark for 6D pose estimation.}
To properly evaluate our vision system independent from the larger robotic system, we have produced a large benchmark dataset with scenarios from APC 2016 with manual labels for objects' segmentation and 6D poses. Previous efforts to construct benchmark datasets include Berkeley's dataset~\cite{singh2014bigbird} with a number of objects from and beyond APC 2015 and Rutgers's dataset~\cite{rennie2016dataset} with semi-automatically labeled data.


\section{Amazon Picking Challenge 2016}


The APC 2016 posed a simplified version of the general picking and stowing tasks in a warehouse. In the picking task, robots sit within a 2x2 meter area in front of a shelf populated with objects, and autonomously pick 12 desired items and place them in a tote. In the stowing task, robots pick all 12 items inside a tote and place them in a pre-populated shelf.

Before the competition, teams were provided with a list of 39 possible objects along with 3D CAD models of the shelf and tote. At run-time, robots were provided with the initial contents of each bin in the shelf and a work-order containing which items to pick. After picking and stowing the appropriate objects, the system had to report the final contents of both shelf and tote. Competition details are in~\cite{apcwebsite}.

\section{System Description}
Our vision system takes in RGB-D images from multiple views, and outputs 6D poses and a segmented point cloud for the robot to complete the picking and stowing tasks. 



The camera is compactly integrated in the end-effector of a 6DOF industrial manipulator ABB IRB1600id, and points at the tip of the fingers (\figref{fig:teaser}). This configuration gives the robot full controllability of the camera viewpoint, and provides feedback about grasp or suction success.
%
%
%
The camera of choice is the RealSense F200 RGB-D because its depth range (0.2m--1.2m) is appropriate for close manipulation, and because it is a consumer-level range sensor with a decent amount of flexibility on the data capture process. 

Due to the tight integration of the camera, the gripper fingers, even when fully open, occupy a small portion of the view frustum. We overcome this limitation by combining different viewpoints, making use of the accurate forward kinematic reported by the robot controller.

\section{6D Object Pose Estimation}
\label{sec:pose_estimate}
The algorithm estimates the 6D pose of all objects in a scene in two phases (Figure \ref{fig:overview}).
First, it segments the RGB-D point clouds captured from multiple views into different objects using a deep convolutional neural network.
Second, it aligns pre-scanned 3D models of the identified objects to the segmented point clouds to estimate the 6D pose of each object.
Our approach is based on well-known methods. However, our evaluations show that when used alone, they are far from sufficient. In this section we present brief descriptions of these methods followed by in-depth discussions of how we combine them into a robust vision system.

\begin{figure}[t]
\vspace{2mm}
\centering
  \includegraphics[width=8.5cm]{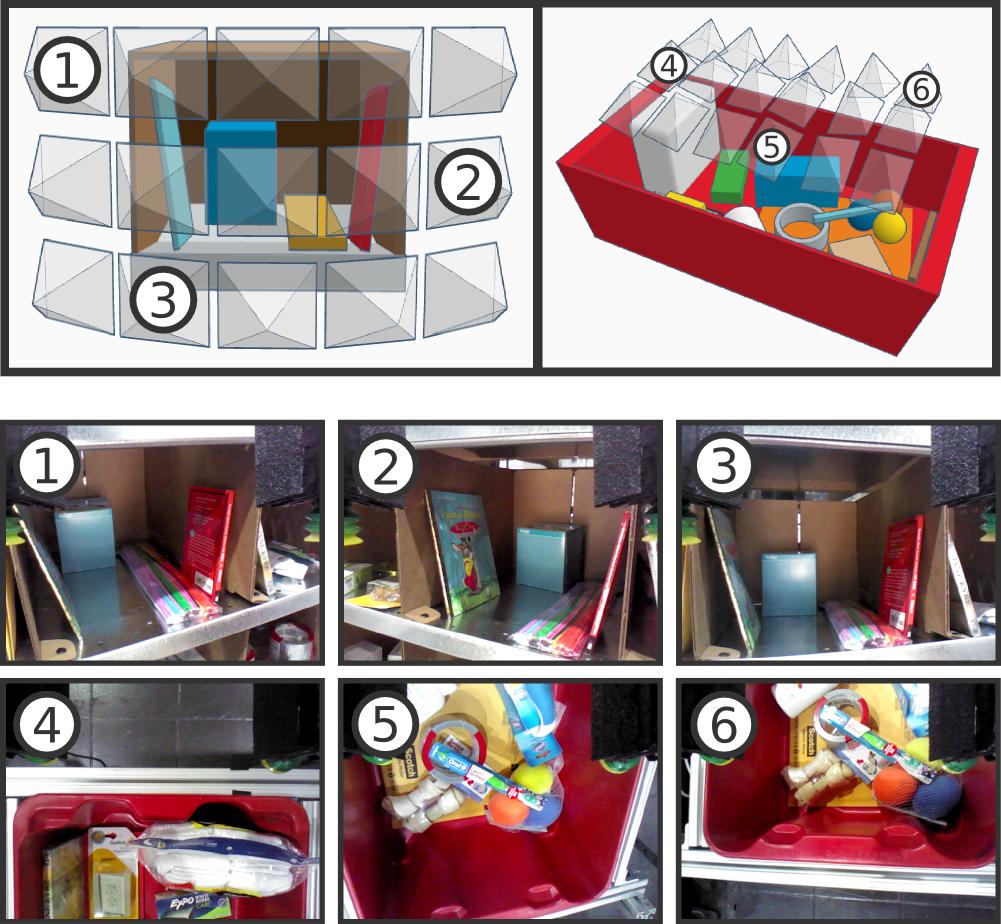}
  \caption{Camera viewpoints of the RGB-D frames captured for bins and tote, and captured color images from 6 selected viewpoints. The 15 viewpoints of a shelf bin (upper-left) are arranged in a 3x5 grid. The 18 viewpoints of a tote (upper-right) are arranged in a 3x6 grid.}
  \label{fig:viewpoint}
  \vspace{-5mm}
\end{figure}

\subsection{Object Segmentation with Fully Convolutional Networks}
In recent years, ConvNets have made tremendous progress for computer vision tasks~\cite{girshick2014rich,long2015fully}. We leverage these advancements to segment camera data into the different objects in the scene. More explicitly, we train a VGG architecture \cite{simonyan2014very} Fully Convolutional Network (FCN) \cite{long2015fully} to perform 2D object segmentation. The FCN takes an RGB image as input and returns a set of 40 densely labeled pixel probability maps--one for each of the 39 objects and one for the background--of the same dimensions as the input image.

\myparagraph{Object segmentation using multiple views.}
Information from a single camera view and from a given object, is often limited due to clutter, self-occlusions, and bad reflections. We address the missing information during the model-fitting phase by combining information from multiple views so that the object surfaces are more distinguishable.
In particular, we feed the RGB images captured from each viewpoint (18 for stowing from the tote and 15 for picking from the shelf) to the trained FCN, which returns a 40 class probability distribution for each pixel in each RGB-D image. 
After filtering by the list of expected objects in the scene, we threshold the probability maps (three standard deviations above the mean probability across all views) and ignore any pixels whose probabilities for all classes are below these thresholds. 
We then project the segmented masks for each object class into 3D space 
and directly combine them into a single segmented point cloud with the forward kinematic feedback from the robot arm (note that segmentation for different object classes can overlap each other). 

\begin{figure}[t]
\vspace{2mm}
\centering
  \includegraphics[width=8.5cm]{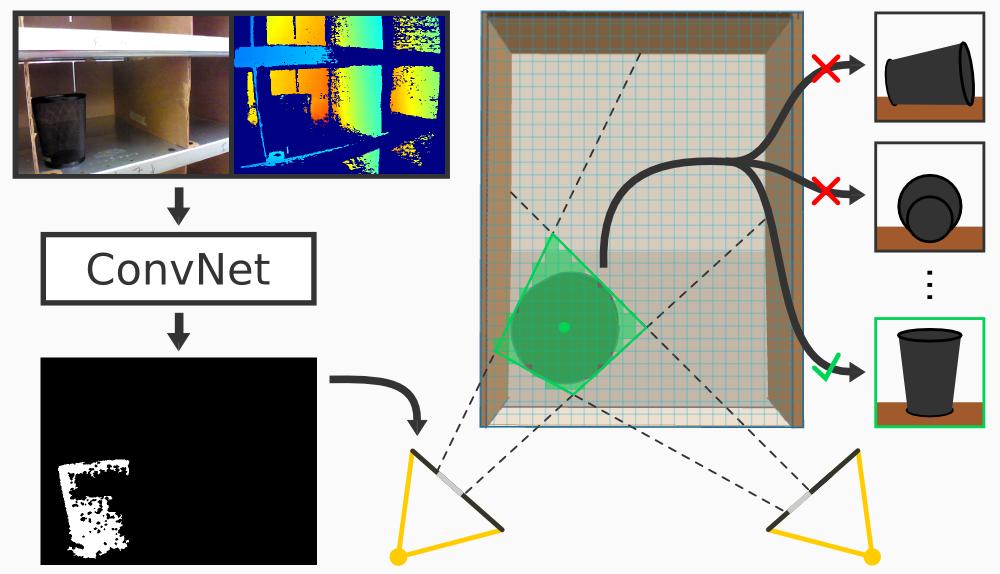}
  \caption{Pose estimation for objects with no depth. 2D object segmentation results from a fully convolutional network are triangulated between the different camera views to create a 3D convex hull (green) of the object. For simplicity, only two camera views (yellow) are illustrated. The centroid and aspect ratio of the convex hull are used to estimate the geometric center of the object and its orientation (from a predefined set of expected orientations).}
\vspace{-5mm}
\end{figure}

\myparagraph{Reduce noise in point cloud.} 
Fitting pre-scanned models to the segmented point cloud directly often gives poor results because of noise from the sensor and noise from the segmentation. We address this issue in three steps:
First, to reduce sensor noise, we eliminate spatial outliers from the segmented point cloud, by removing all point farther than a threshold from its k-nearest neighbors.
Second, to reduce segmentation noise, especially on the object boundaries, we remove points that lie outside the shelf bin or tote, and those that are close to a pre-scanned background model.
Finally, we further filter outlier points from each segmented group of points by finding the largest contiguous set of points along each principal axis (computed via PCA) and remove any points that are disjoint from this set.

\myparagraph{Handle object duplicates.} 
Warehouse shelves commonly contain multiple instances of the same object. Naively segmenting RGB-D data will treat two distinct object with the same label as the same object. Since we know the inventory list in the warehouse setting, we know the number of identical objects we expect in the scene. 
We make use of k-means clustering to separate the segmented and aggregated point cloud into the appropriate number of objects. Each cluster is then treated independently during the model-fitting phase of the algorithm.

\subsection{3D Model-Fitting}
We use the iterative closest point (ICP) algorithm  \cite{gelfand2003geometrically} on the segmented point cloud to fit pre-scanned 3D models of objects and estimate their poses. The vanilla ICP algorithm, however, gives nonsensical results in many scenarios. We describe here several such pitfalls along with our solutions.

\myparagraph{Point clouds with non-uniform density.}
In a typical RGB-D point cloud, surfaces perpendicular to the sensor's optical axis have often denser point clouds. The color of the surface changes its reflectivity on the IR spectrum, which also affects the effective point cloud density.
These non-uniformities are detrimental to the ICP algorithm because it biases convergence toward denser areas.  By applying a 3D uniform average grid filter to the point clouds, we are able to give them consistent distributions in 3D space.

\myparagraph{Pose initialization.} 
ICP is an iterative local optimizer, and as such, it is sensitive to initialization.
The principal directions of the segmented point cloud, as estimated by PCA, give us a reasonable first approximation to the orientation of objects with uneven aspect ratios.
We have observed experimentally that the choice of initial orientation for objects with even aspect ratios has little effect on the final result of ICP.
Analogously, one would use the centroid of the point cloud as the initial guess for the geometric center of the object, however we have observed that since captured point clouds are only partial, those two centers are usually biased from each other.
To address this, we push back the initial pose of the pre-scanned object back along the optical axis of the RGB-D camera by half the size of the object's bounding box, under the naive assumption that we are only seeing ``half" the object. This initialization has proven more successful in avoiding local optimums.

\begin{figure}[t]
\vspace{2mm}
\centering
  \includegraphics[width=8.3cm]{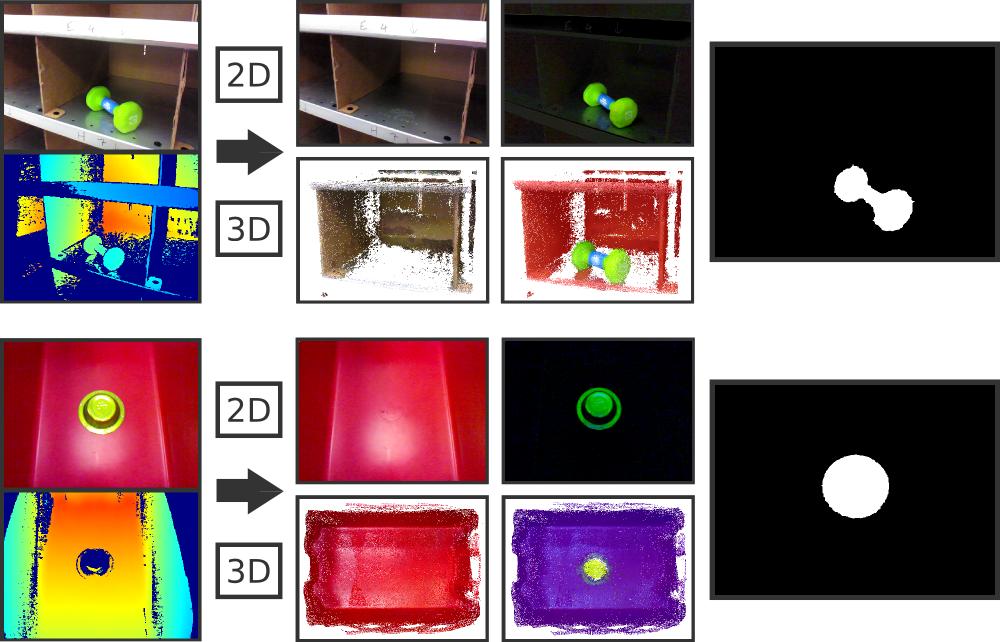}
  \vspace{-1mm}
  \caption{To automatically obtain pixel-wise object labels, we separate the target objects from the background to create an object mask. There are a 2D and a 3D component in this data process. Both use color and depth information. The 2D pipeline is robust to thin objects and objects with no depth, while the 3D pipeline is robust to an unstable background. 
  }
\vspace{-4mm}
\end{figure}

\myparagraph{Coarse to fine ICP.}
Even after reducing noise in the segmentation step, the resulting point cloud may still have noise (e.g., mislabeled points from adjacent objects). We address this with two passes of ICP, acting on different subsets of the point cloud: we define the \emph{inlier} threshold of an ICP iteration as the percentile L2 distance above which we ignore. ICP with a 90\% inlier ratio keeps the closest pairs of points between the two point clouds up to the 90th percentile.
The main assumption is that regions of the point cloud that are correctly labeled are denser than regions with incorrect label. A first pass with a high inlier threshold (90\%) moves the pre-scanned complete model closer to the correct portion of the partial view than the noisy portion. 
Starting now from a coarse but robust initialization, the second pass uses a lower inlier threshold (45\%) to ignore the noisy portion of the point cloud and converge to a more accurate pose.

\subsection{Handling Objects with Missing Depth.}
Many objects in the APC, as it is typical in retail warehouses, have surfaces that challenge infrared-based depth sensors, e.g., with plastic wrapping returning noisy or multiple reflections, or transparent or meshed materials which may not register at all. For these objects the captured point cloud is noisy and sparse, and our pose estimation algorithm performs poorly.

Our solution leverages the multi-view segmentation to estimate a convex hull of the object by carving a 3D gridded space of voxels with the segmented RGB images. This process results in a 3D mask that encapsulates the real object. We use the convex hull of that mask to estimate the geometric center of the object and approximate its orientation (assuming that the object is axis-aligned).


\section{Self-supervised Training}
By bringing deep learning into the approach we gain robustness. This, however, comes at the expense of amassing quality training data, which is necessary to learn high-capacity models with many parameters.
Gathering and manually labeling such large amounts of training data is expensive. The existing large-scale datasets used by deep learning (e.g. ImageNet~\cite{ILSVRC15}) are mostly Internet photos, which have very different object and image statistics from our warehouse setting.

To automatically capture and pixel-wise label images, we propose a self-supervised method, based on three observations: 
\begin{itemize}
\item[$\Cdot$] Batch-training on scenes with a single object can yield deep models that perform well on scenes with multiple objects~\cite{girshick2014rich} (i.e., simultaneous training on cat-only or dog-only images enables successful testing on cat-with-dog images); 
\item[$\Cdot$] An accurate robot arm and accurate camera calibration, gives us at will control over camera viewpoint;
\item[$\Cdot$] For single object scenes, with known background and known camera viewpoint, we can automatically obtain precise segmentation labels by foreground masking.
\end{itemize}
The captured training dataset contains 136,575 RGB-D images of 39 objects, all automatically labeled.

\myparagraph{Semi-automatic data gathering.} To semi-autonomously gather large quantities of training data, we place single known objects inside the shelf bins or tote in arbitrary poses, and configure the robot to move the camera and capture RGB-D images of the objects from a variety of different viewpoints.
The position of the shelf/tote is known to the robot, as is the camera viewpoint, which we use to transform the collected RGB-D images in shelf/or tote frame.
After capturing several hundred RGB-D images, the objects are manually re-arranged to different poses, and the process is repeated several times. 
Human involvement sums up to re-arranging the objects and labeling which objects correspond to which bin/tote. 
Selecting and changing the viewpoint, capturing sensor data, and labeling each image by object is automated.
We collect RGB-D images of the empty shelf and tote from the same exact camera viewpoints to model the background, in preparation for the automatic data labeling.

\myparagraph{Automatic data labeling.} To obtain pixel-wise object segmentation labels, we create an object mask that separates foreground from background. The process is composed of 2D and 3D pipelines. The 2D pipeline is robust to thin objects (objects not sufficient volume to be reliably segmented in 3D when placed too close to a walls or ground) and objects with no depth information, while the 3D pipeline is robust to large miss-alignments between the pre-scanned shelf bin and tote. Results from both pipelines are combined to automatically label an object mask for each training RGB-D image.

The 2D pipeline starts by fixing minor possible image misalignments by using multimodal 2D intensity-based registration to align the two RGB-D images \cite{styner2000parametric}. We then convert the aligned color image from RGB to HSV, and do pixel-wise comparisons of the HSV and depth channels to separate and label foreground from background.

The 3D pipeline uses multiple views of an empty shelf bin and tote to create their pre-scanned 3D models. We then use ICP to align all training images to the background model, and remove points too close to the background to identify the foreground mask. Finally, we project the foreground points back to 2D to retrieve the object mask.

\begin{figure}[t]
\vspace{2mm}
\centering
  \includegraphics[width=8.5cm]{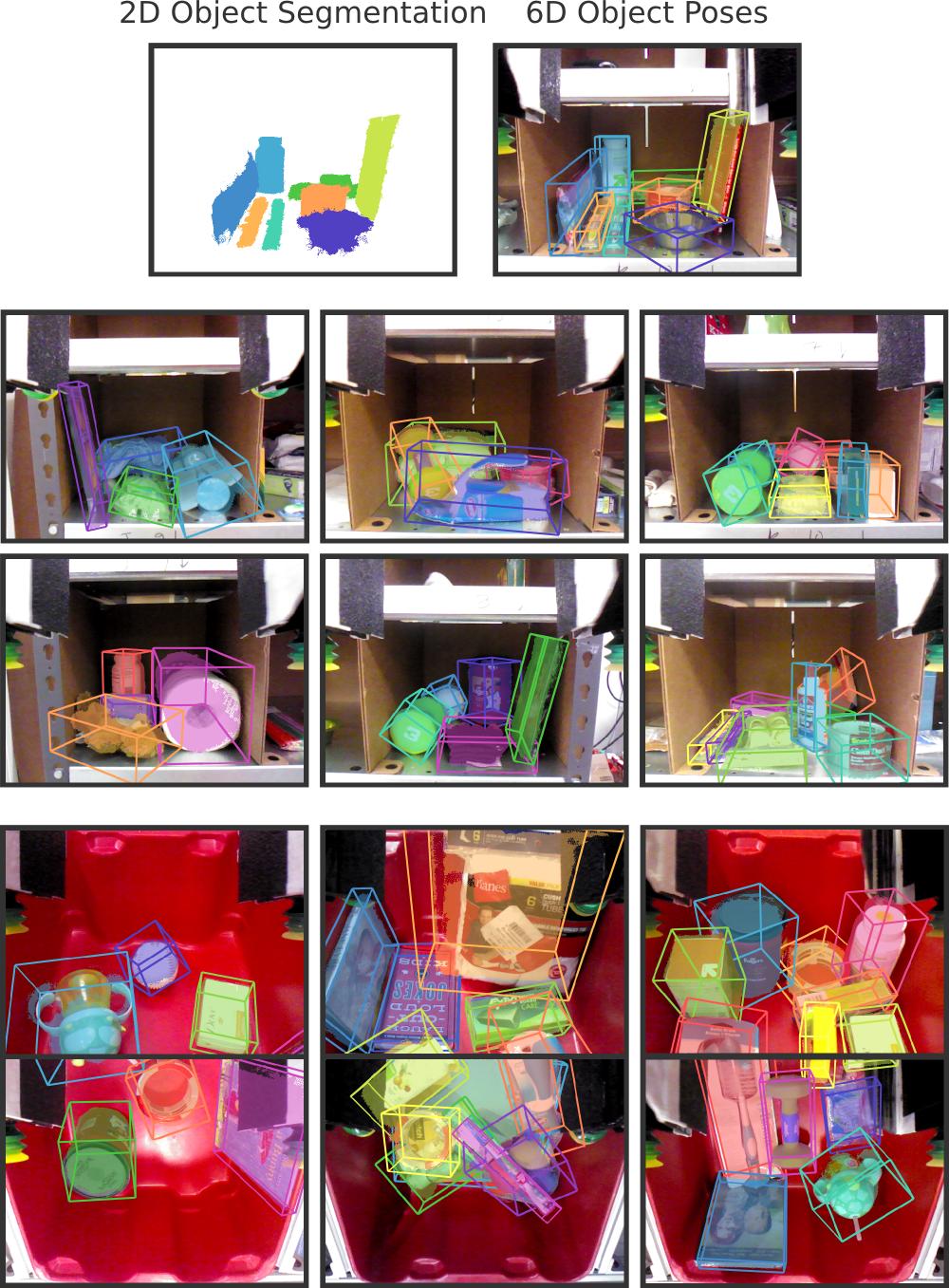}
  \caption{Examples from our benchmark dataset. The dataset contains 477 scenes with 2,087 unique object poses seen from multiple viewpoints. In total, there are 7,713 images with manually-annotated ground truth 6D object poses and segmentation labels.}
\label{fig:benchmark}
\vspace{-5mm}
\end{figure}

\myparagraph{Training neural network.} To leverage features trained from a larger image domain, we use the sizable FCN-VGG network architecture from \cite{simonyan2014very} and initialize the network weights using a model pre-trained on ImageNet for 1000-way object classification. We fine-tune the network over the 40-class output classifier (39 classes for each APC object and 1 class for background) using stochastic gradient descent with momentum. Due to illumination and object viewpoint biases, we maximize performance by training two such segmentation networks: one for shelf bins and one for the tote. The segmentation labels automatically generated for the training data can be noisy. However, we find that the networks are still capable of working well during test time due to the sheer size of available training data.

\begin{figure}[t]
\vspace{2mm}
\centering
  \includegraphics[width=8.5cm]{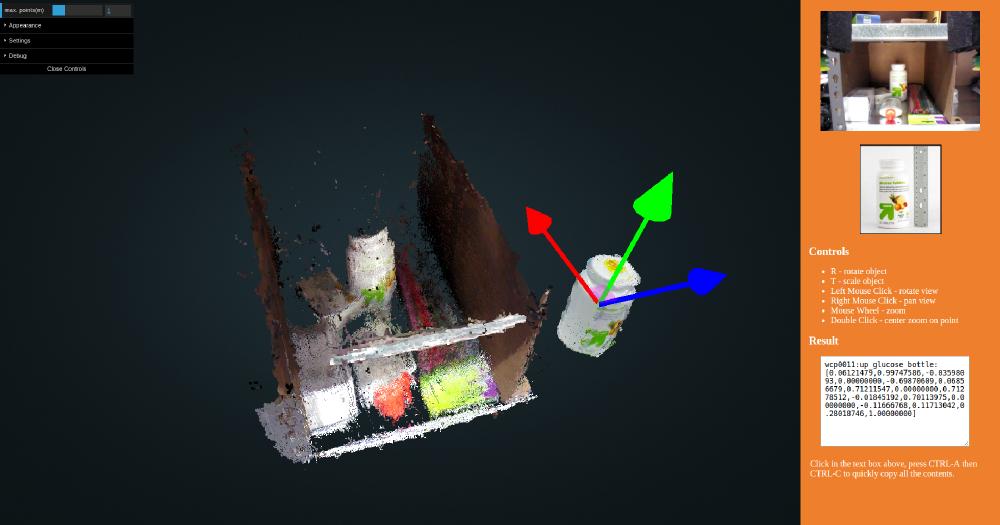}
  \caption{The 3D online annotation tool used to label the benchmark. The drag-and-drop UI allows annotators to navigate in 3D space and manipulate point clouds with ease. Annotators are instructed to move and rotate a pre-scanned object model to its ground truth location in a 3D point cloud generated from RGB-D data. Labeling one object takes about 1 minute.}
  \label{fig:tool}
  \vspace{-5mm}
\end{figure}

\section{Implementation} 
All components of the vision system are modularized into reusable ROS packages, with CUDA GPU acceleration. Deep models are trained and tested with Marvin \cite{Marvin20151110}, a ROS-compatible and lightweight deep learning framework. Training our models takes up to 16 hours prior to convergence. 

Our robot is controlled by a computer with an Intel E3-1241 CPU 3.5 GHz and an NVIDIA GTX 1080. The run-time speeds per component are as follows: 10ms for ROS communication overhead, 400ms per forward pass of VGG-FCN, 1200ms for denoising per scene, and 800ms on model-fitting per object. On average, pose estimation time is 3-5 seconds per shelf bin and 8-15 seconds for the tote. Combined with multi-view robot motions, total visual perception time is 10-15 seconds per shelf bin and 15-20 seconds for the tote.

\section{Evaluation}
We evaluate variants of our method in different scenarios in the benchmark dataset to understand (1) how segmentation performs under different input modalities and training dataset sizes and (2) how the full vision system performs.


\subsection{Benchmark Dataset}
\label{ref:benchmark}
Our benchmark dataset, `Shelf\&Tote', contains over 7,000 RGB-D images spanning 477 (\figref{fig:benchmark}) scenes at $640\times{480}$ resolution. We collected the data during practice runs and competition finals for the APC and manually labeled 6D object poses and segmentations using our online annotator (\figref{fig:tool}). The data reflects various challenges found in the warehouse setting: reflective materials, variation in lighting conditions, partial views, and sensor limitations (noisy and missing depth) over cluttered environments.


Tables \ref{table:seg} and \ref{table:pose} summarize our experimental results and highlight the differences in performance over different overlapping scene categories:
\begin{itemize}
\item[$\Cdot$]\textbf{cptn:} during competition at the APC finals.
\item[$\Cdot$]\textbf{environment:} in an office (off); in the APC competition warehouse (whs).
\item[$\Cdot$]\textbf{task:} \textit{picking} from a shelf bin or \textit{stowing} from a tote.
\item[$\Cdot$]\textbf{clutter:} with multiple objects.
\item[$\Cdot$]\textbf{occlusion:} with \% of object occluded by another object, computed from ground truth.
\item[$\Cdot$]\textbf{object properties:} with objects that are deformable, thin, or have no depth from the RealSense F200 camera.
\end{itemize}

\begin{figure*}[t]
 \vspace{2mm}
\centering
  \includegraphics[width=\textwidth]{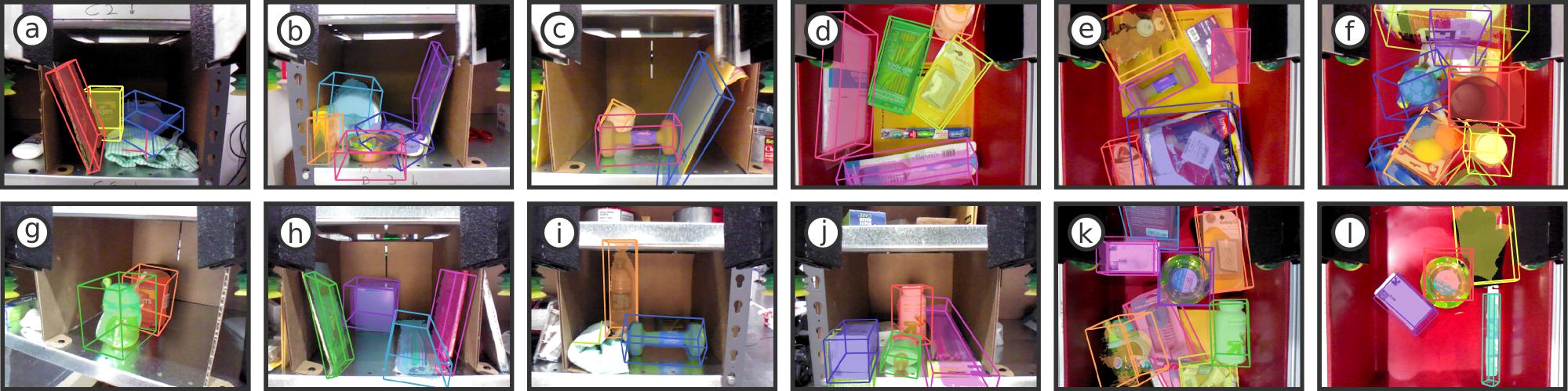}%
    \vspace{-2mm}
  \caption{Example results from our vision system. 6D pose predictions are highlighted with a 3D bounding box. For deformable objects (cloth in a,c,i) we output the center of mass. We additionally illustrate successful pose predictions for objects with missing depth (water bottle, black bin, green sippy cup, green bowl)}
  \label{fig:successful_results}

\vspace{2mm}
\centering
  \includegraphics[width=\textwidth]{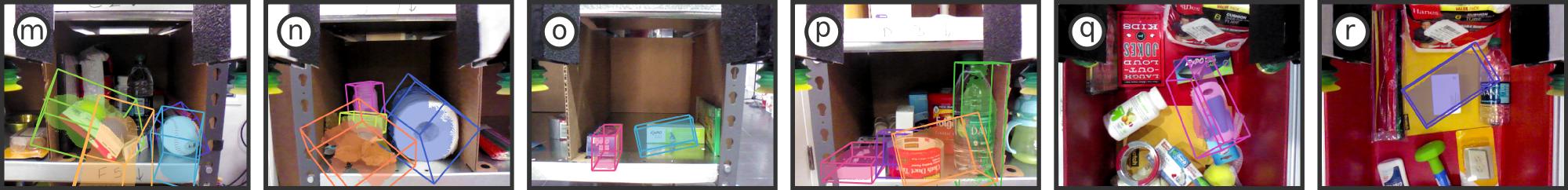}%
  \vspace{-2mm}
  \caption{Several common failure cases. These include low-confidence predictions due to severe occlusion (missing object labels in m,o,p), completely incorrect pose predictions due to confusion in texture (m,p,r) or bad initialization (n,q), and model-fitting errors (o).}
  \label{fig:failure_results}
  \vspace{-5mm}
\end{figure*}

\subsection{Evaluating Object Segmentation}

We test several variants of our FCN on object segmentation to answer two questions: (1) can we leverage both color and depth segmentation? (2) is more training data useful?

\myparagraph{Metrics.} We compare the predicted object segmentation from our trained FCNs against the ground truth segmentation labels of the benchmark dataset using pixel-wise precision and recall. Table \ref{table:seg} displays the mean average F-scores ($F=2\cdot\frac{\text{precision}\cdot\text{recall}}{\text{precision}+\text{recall}}$).

\myparagraph{Depth for segmentation.} We use HHA features \cite{gupta2014learning} to encode depth information into three channels: horizontal disparity, height above ground, and angle of local surface normal with the inferred direction of gravity. We compare AlexNet trained on this encoding, VGG on RGB data, and both networks concatenated in Table \ref{table:seg}.

We find that adding depth does not yield any notable improvements in segmentation performance, which could be in part due to the noisiness of the depth information from our sensor. On the other hand, we observe that the FCN performs significantly better when trained on color data, with the largest disparity for deformable objects and thin objects, whose textures provide more discriminative power than their geometric structure.

\myparagraph{Size of training data.} Deep learning models have seen significant success, especially if given large amounts of training data. However in our scenario---instance-level object segmentation on few object categories---it is not clear whether such a large dataset is necessary. We create two new datasets by randomly sampling 1\% and 10\% of the original and use them to train two VGG FCNs (Table \ref{table:seg}). We confirm marked improvements in F-score across all benchmark categories going from 1\% to 10\% to 100\% of training data.




\subsection{Evaluating Pose Estimation}

We evaluate several key components of our vision system to determine whether they increase performance in isolation.

\myparagraph{Metrics.} We report the percentage of object pose predictions with error in orientation smaller than 15$^{\circ}$, and the percentage with error in translation smaller than 5cm. The metric also recognizes the structural invariance of several objects, some of which are axially-symmetric (cuboids), radially-symmetric (bottles, cylinders), or deformable (see web page \cite{projectwebsite} for further details). We have observed experimentally that these bonds of 15$^{\circ}$ and 5cm are sufficient for picking with sensor-guarded motions.

\myparagraph{Multi-view information.} With multiple views the system overcomes missing information due to self-occlusions,  other-object occlusions, or clutter. Multi-view information also alleviates problems with illumination on reflective surfaces. 

To quantify the effect of the multiple-view system, we test the full vision system on the benchmark with three different subsets of camera views: 
\begin{itemize}
\item[$\Cdot$] [Full] All 15 views for shelf bins $a_{1_\text{shelf}}$ = \{0 \ldots 14\} and all 18 views for the tote $a_{1_\text{tote}}$ = \{0 \ldots 17\}.
\item[$\Cdot$] [5v-10v] 5 views for shelf $a_{2_\text{shelf}}$ = \{0,4,7,10,14\} and 10 views for tote $a_{2_\text{tote}}=$\{0,2,4,6,8,9,11,13,15,17\}, with a sparse arrangement and a preference for wide-baseline view angles.
\item[$\Cdot$] [1v-2v] 1 view for shelf bins $a_{3_\text{shelf}}$ = \{7\} 
and 2 views for the tote $a_{3_\text{tote}}$ =\{7,13\}.
\end{itemize}
The viewpoint ids are zero-indexed in row-major order as depicted in Figure \ref{fig:viewpoint}.
Our results show that multiple views robustly address occlusion and heavy clutter in the warehouse setting (Table \ref{table:pose} [clutter] and [occlusion]).
They also present a clear contrast between the performance of our algorithm using a single view of the scene, versus multiple views of the scene (Table \ref{table:pose} [Full] v.s [1v-2v]). 


\myparagraph{Denoising.} The denoising step described in \secref{sec:pose_estimate} proves important for achieving good results. With this turned off, the accuracy in estimating the translation and rotation decreases by 6.0\% and 4.4\% respectively (Table \ref{table:pose}).

\myparagraph{ICP improvements.} Without the pre-processing steps to ICP, we observe a drop in prediction accuracy of 0.9\% in translation and 3.1\% in rotation (Table \ref{table:pose}).

\myparagraph{Performance upper bound.} We also evaluated how well the model-fitting part of our algorithm alone performs on the benchmark by using ground truth segmentation labels from the benchmark as the performance upper bound.

\subsection{Common Failure Modes} 
Here we summarize the most common failure modes of our vision system, which are illustrated in Figure \ref{fig:failure_results}: 
\begin{itemize}
    \item[$\Cdot$] The FCN segmentation for objects under heavy occlusion or clutter are likely to be incomplete resulting in poor pose estimation (Fig.~\ref{fig:successful_results}.e), or undetected (Fig.~\ref{fig:failure_results}.m and p). This happens with more frequency at back of the bin with poor illumination. 
    \item[$\Cdot$] Objects color textures are confused with each other. Figure \ref{fig:failure_results}.r shows a Dove bar (white box) on top of a yellow Scotch mail envelope, which combined have a similar appearance to the outlet plugs.
    \item[$\Cdot$] Model fitting for cuboid objects often confuses corner alignments (marker boxes in Fig.~\ref{fig:failure_results}.o). This inaccuracy, however, is still within the range of tolerance that the robot can tolerate thanks to sensor-guarded motions.
\end{itemize}



\myparagraph{Filtering failure modes by confidence score.} 
We compute a confidence score per object pose prediction that favors high precision for low recall. 
Specifically, the confidence score of a pose prediction equals the mean value of confidence scores over all points belonging to the segmentation of the object. 
We observe that erroneous poses (especially those due to partial occlusions) more often have low confidence scores.
The robot system uses this value to target only predictions with high scores.

We evaluate the usefulness of the confidence scores by recalling the output of the perception system to only consider predictions with confidence scores larger than $10\%$ and $70\%$ respectively (see Table \ref{table:pose}). 
These confidence percentages are important thresholds, because the full robot system, predictions with $< 10\%$ confidence (conf-10, at 78\% recall) are ignored during planning, and prediction with $> 70\%$ confidence (conf-70, at 23\% recall) trigger a pick attempt.

\begin{table*}[t]
\vspace{2mm}
  \centering
  \caption{2D object segmentation evaluation (pixel-level object classification average \% F-scores).}
  \vspace{-2mm}
  \begin{tabular}{c|c|c|c|c|c|c|c|c|c|c|c|c|c|c|c}
    \hline
    \multicolumn{3}{c}{ } & \multicolumn{2}{|c}{environment} & \multicolumn{2}{|c}{task} & \multicolumn{3}{|c}{clutter (\# of objects)} & \multicolumn{3}{|c}{occlusion (\%)} & \multicolumn{3}{|c}{object-specific properties} \\\hline
    network & all & cptn & off & whs & shelf & tote & 1 - 3 & 4 - 5 & 6 + & $<$ 5 & 5 - 30 & 30 + & dfrm. & no depth & thin \\\hline
    color & \bf 45.5 & \bf 42.7 & \bf 46.8 & \bf 44.2 & \bf 47.7 & \bf 43.7 & \bf 53.0 & \bf 46.0 & \bf 42.2 & \bf 49.9 & \bf 41.4 & \bf 33.3 & \bf 54.0 & \bf 47.9 & \bf 41.7 \\ 
    color+depth & 43.8 & 41.5 & 44.8 & 42.6 & 45.8 & 41.9 & 52.2 & 43.5 & 40.0 & 47.5 & 39.1 & 32.6 & 51.1 & 47.7 & 37.2 \\
    depth & 37.1 & 35.0 & 38.6 & 35.5 & 39.8 & 34.9 & 45.5 & 37.0 & 33.5 & 40.8 & 33.2 & 26.3 & 44.1 & 42.3 & 29.1\\\hline
    10\% data & 20.4 & 18.8 & 19.5 & 21.3 & 21.7 & 20.3 & 36.0 & 21.6 & 18.0 & 21.2 & 25.5 & 0.0 & 41.9 & 17.2 & 33.3\\
    1\% data & 8.0 & 9.2 & 7.2 & 8.8 & 15.8 & 6.5 & 17.3 & 7.5 & 6.0 & 7.7 & 8.3 & 7.8 & 10.1 & 5.7 & 3.5\\\hline
  \end{tabular}
  \label{table:seg}
\end{table*}

\begin{table*}[t]
\vspace{-2mm}
  \centering
  \caption{Full vision system evaluation (average \% correct rotation and translation predictions for object pose)}
  \vspace{-2mm}
  \begin{tabular}{c|c|c|c|c|c|c|c|c|c|c|c|c|c|c|c}
    \hline
    \multicolumn{3}{c}{ } & \multicolumn{2}{|c}{environment} & \multicolumn{2}{|c}{task} & \multicolumn{3}{|c}{clutter (\# of objects)} & \multicolumn{3}{|c}{occlusion (\%)} & \multicolumn{3}{|c}{object-specific properties} \\\hline
    algorithm & all & cptn & off & whs & shelf & tote & 1 - 3 & 4 - 5 & 6 + & $<$ 5 & 5 - 30 & 30 + & dfrm. & no depth & thin \\\hline
    Full (rot.) & \bf 49.8 & \bf 62.9 & \bf 52.5 & \bf 47.1 & \bf 50.4 & \bf 49.3 & \bf 56.1 & \bf 54.6 & \bf 45.4 & \bf 56.9 & \bf 43.2 & \bf 33.9 & \bf - & \bf 55.6 & \bf 54.7\\
    Full (trans.) & \bf 66.1 & \bf 71.0 & \bf 66.3 & \bf 65.9 & \bf 63.4 & \bf 68.1 & \bf 76.7 & \bf 66.7 & \bf 61.9 & \bf 79.4 & \bf 57.4 & \bf 27.3 & \bf 75.4 & \bf 63.3 & \bf 58.1\\\hline\hline
    5v-10v (rot.) & 44.0 & 48.6 & 50.9 & 35.9 & 50.9 & 38.9 & 53.9 & 53.1 & 34.4 & 47.6 & 40.0 & 26.7 & - & 47.4 & 42.4 \\
    5v-10v (trans.) & 58.4 & 50.0 & 63.7 & 52.1 & 61.0 & 56.5 & 69.4 & 63.0 & 50.3 & 66.2 & 49.8 & 21.3 & 54.7 & 67.3 & 35.4 \\\hline
    1v-2v (rot.) & 38.9 & 60.0 & 41.1 & 36.5 & 45.0 & 35.3 & 45.7 & 45.2 & 32.7 & 43.6 & 33.9 & 14.8 & - & 40.9 & 35.4 \\
    1v-2v (trans.) & 52.5 & 50.0 & 56.3 & 48.2 & 53.8 & 51.8 & 60.4 & 56.5 & 46.7 & 58.2 & 47.8 & 16.7 & 52.9 & 55.9 & 33.3 \\\hline\hline
    
    conf-70 (rot.) & 58.3 & 77.3 & 65.0 & 49.0 & 64.2 & 53.2 & 63.8 & 69.3 & 49.0 & 63.7 & 43.1 & 36.4 & - & 64.5 & 81.6\\
    conf-70 (trans.) & 84.5 & 95.5 & 84.7 & 84.2 & 82.6 & 86.1 & 86.2 & 84.1 & 83.2 & 87.1 & 77.1 & 72.7 & 83.1 & 77.4 & 85.7\\\hline
    conf-10 (rot.) & 55.0 & 70.8 & 57.0 & 52.7 & 54.9 & 55.0 & 58.6 & 59.3 & 51.0 & 59.8 & 50.0 & 34.2 & - & 53.1 & 60.2\\
    conf-10 (trans.) & 76.5 & 81.2 & 76.7 & 76.3 & 73.4 & 79.1 & 80.8 & 74.4 & 75.4 & 84.0 & 70.0 & 40.0 & 78.1 & 72.0 & 70.1\\\hline\hline
    
    no denoise (rot.) & 43.8 & 45.6 & 46.9 & 40.6 & 45.3 & 42.7 & 52.0 & 46.7 & 39.5 & 51.1 & 37.3 & 28.1 & - & 48.8 & 54.1\\
    no denoise (trans.) & 61.7 & 66.4 & 61.9 & 61.5 & 60.4 & 62.6 & 74.8 & 62.7 & 56.4 & 76.5 & 52.9 & 19.9 & 75.0 & 62.3 & 53.8\\\hline
    no ICP+ (rot.) & 48.9 & 60.8 & 51.2 & 46.7 & 49.1 & 48.8 & 55.4 & 54.1 & 44.4 & 55.8 & 41.9 & 36.2 & - & 53.6 & 52.5\\
    no ICP+ (trans.) & 63.0 & 67.2 & 63.2 & 62.9 & 59.7 & 65.4 & 72.1 & 64.4 & 59.1 & 75.2 & 57.0 & 24.6 & 67.3 & 62.8 & 53.2\\\hline\hline
    
    gt seg rot. & 63.4 & 74.4 & 65.8 & 60.9 & 68.1 & 60.1 & 69.1 & 68.8 & 59.1 & 67.6 & 60.0 & 53.5 & - & 58.0 & 74.1 \\
    gt seg trans. & 88.1 & 90.4 & 85.7 & 90.4 & 86.9 & 88.9 & 88.3 & 88.0 & 88.0 & 90.7 & 90.3 & 71.4 & 90.5 & 71.5 & 79.8\\\hline
  \end{tabular}
  \label{table:pose}
  \vspace{-5mm}
\end{table*}

\section{Discussion}
Despite tremendous advances in computer vision, many state-of-the-art well-known approaches are often insufficient for relatively common scenarios. We describe here two observations that can lead to improvements in real systems:

\myparagraph{Make the most out of every constraint.}
External constraints limit what systems can do. Indirectly they also limit the set of states in which the system can be, which can lead to opportunities for simplifications and robustness in the perception system.
In the picking task, each team received a list of items, their bin assignments, and a model of the %
shelf. 
All teams used the bin assignments to rule out objects from consideration and the model of the shelf to calibrate their robots. These optimizations are straightforward and useful.
However, further investigation yields more opportunity. By using these same constraints, we constructed a self-supervising mechanism to train a deep neural network with significantly more data. As our evaluations show, the volume of training data is strongly correlated with performance.

\myparagraph{Designing robotic and vision systems hand-in-hand.}
Vision algorithms are too often designed in isolation.
However, vision is one component of a larger robotic system with needs and opportunities.
Typical computer vision algorithms operate on single images for segmentation and recognition. Robotic arms free us from that constraint, allowing us to precisely fuse multiple views and improve performance in cluttered environments.
Computer vision systems also tend to have fixed outputs (e.g., bounding boxes or 2D segmentation maps), but robotic systems with multiple manipulation strategies can benefit from variety in output. For example, suction cups and grippers might have different perceptual requirements. While the former might work more robustly with a segmented point cloud, the latter often requires knowledge of the object pose and geometry.

\section{Conclusion} %
In this paper, we present the vision system of Team MIT-Princeton's 3rd- and 4th-place entry in the 2016 Amazon Picking Challenge. To address the challenges posed by the warehouse setting, our framework leverages multi-view RGB-D data and data-driven, self-supervised deep learning to reliably estimate the 6D poses of objects under a variety of scenarios. We also provide a well-labeled benchmark dataset of APC 2016 containing over 7,000 images from 477 scenes.
\bibliographystyle{IEEEtranN} 
{\footnotesize \bibliography{main}} 

\end{document}